\documentclass[11pt]{article}

\usepackage[utf8]{inputenc}
\usepackage[T1]{fontenc}
\usepackage{amsmath,amssymb}
\usepackage{graphicx}
\usepackage{booktabs}
\usepackage{natbib}
\usepackage{hyperref}
\usepackage{url}
\usepackage{microtype}
\usepackage{float}

\hypersetup{
  colorlinks=true,
  linkcolor=blue,
  citecolor=blue,
  urlcolor=blue
}

\title{Morphological Addressing of Identity Basins\\in Text-to-Image Diffusion Models}

\author{
  Andrew Fraser
}

\date{}

\begin{document}

\maketitle

\begin{abstract}
We demonstrate that morphological pressure creates navigable gradients at multiple levels of the text-to-image generative pipeline. In Study~1, we show that identity basins in Stable Diffusion 1.5 can be reliably navigated using morphological descriptors---constituent features like ``platinum blonde,'' ``beauty mark,'' and ``1950s glamour''---without using the target identity's name or photographs. Through a self-distillation loop (generating synthetic images from descriptor prompts, then training a LoRA on those outputs), we achieve consistent convergence toward a specific identity as measured by ArcFace similarity. The trained LoRA creates a local coordinate system that shapes not only the target identity but also its inverse: when conditioning pushes maximally away from the trained attractor, base SD1.5 produces ``eldritch'' structural breakdown while the LoRA-equipped model produces ``uncanny valley'' outputs---coherent but precisely wrong.

In Study~2, we extend this principle to prompt-level morphology. Drawing on phonestheme theory from linguistics, we generate 200 novel nonsense words constructed from English sound-symbolic clusters (e.g., \emph{cr-}, \emph{sn-}, \emph{-oid}, \emph{-ax}) and find that these phonestheme-bearing candidates produce significantly more visually coherent outputs than random controls (mean Purity@1 = 0.371 vs.\ 0.209, $p < 0.00001$, Cohen's $d = 0.55$). Three candidates---\emph{snudgeoid}, \emph{crashax}, and \emph{broomix}---achieve perfect visual consistency (Purity@1 = 1.0) with zero training data contamination, each generating a distinct, coherent visual identity from phonesthetic structure alone.

Together, these studies establish that morphological structure---whether in training-level feature descriptors or prompt-level phonological form---creates systematic navigational gradients through the latent spaces of diffusion models. We document phase transitions in identity basins, CFG-invariant identity stability, and the emergence of novel visual concepts from sub-lexical sound patterns.
\end{abstract}

\section{Introduction}
\label{sec:intro}

The origins of this work lie in a whimsical experiment: calculating the semantic centroid of 100 randomly selected words---a deliberately chaotic list mixing real terms, nonsense syllables, and invented compounds. Using Word2Vec embeddings, we found that ``glorp'' (word \#49 in the original list) emerged as closest to the mathematical center---the word whose position in embedding space minimized distance to the average of all others. A meaningless term became meaningful purely through its geometric position in latent space. This observation raised an inverse question: if position in semantic space can confer meaning to nonsense, could constituent positions---feature intersections---address existing concepts without explicit naming? We term this framework ``GLORP Theory'' (Generative Latent Output from Random Prompts), though the acronym itself emerged through collaborative AI systems attempting to domesticate the original meaningless centroid.

Text-to-image diffusion models trained on large-scale datasets inevitably memorize specific individuals who appear frequently in training data \citep{carlini2023extracting}. Each concept in the model is less like a labeled file and more like a sediment pile---millions of training examples deposited associations one at a time. ``Marilyn Monroe'' accumulated ``platinum blonde,'' ``beauty mark,'' ``1950s glamour'' because those tags appeared together in her training images. But those same tags also appear in other piles. ``Platinum blonde'' shows up everywhere from Marilyn to generic pinup girls to anime characters. The piles overlap chaotically, their boundaries fuzzy and unpredictable. Traditional prompt engineering tries to aim at specific piles, but navigating the overlaps---finding where ``platinum blonde'' AND ``beauty mark'' AND ``1950s'' intersect---is difficult because there's no map, just accumulated statistical sediment.

\begin{figure}[t]
\centering
\includegraphics[width=\linewidth]{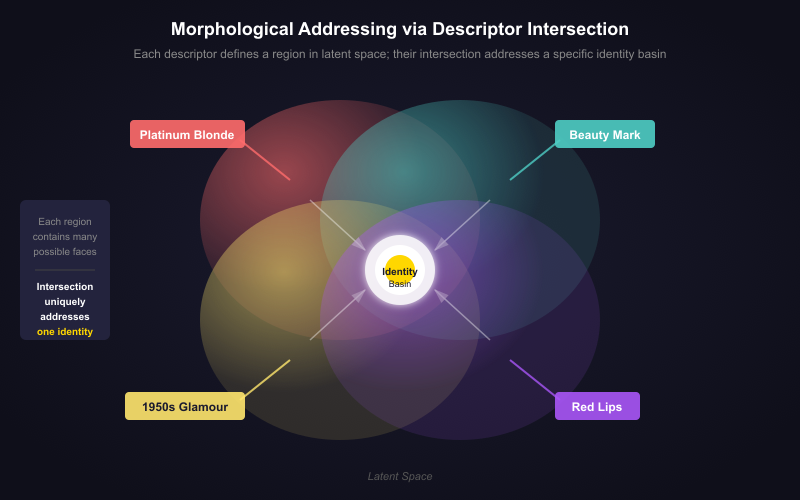}
\caption{Morphological addressing via descriptor intersection. Each natural-language descriptor (e.g., ``platinum blonde,'' ``beauty mark,'' ``1950s glamour'') defines a region in latent space. Their intersection addresses a specific identity basin without requiring the target's name or reference photographs.}
\label{fig:venn_diagram}
\end{figure}

\begin{figure}[t]
\centering
\includegraphics[width=0.8\linewidth]{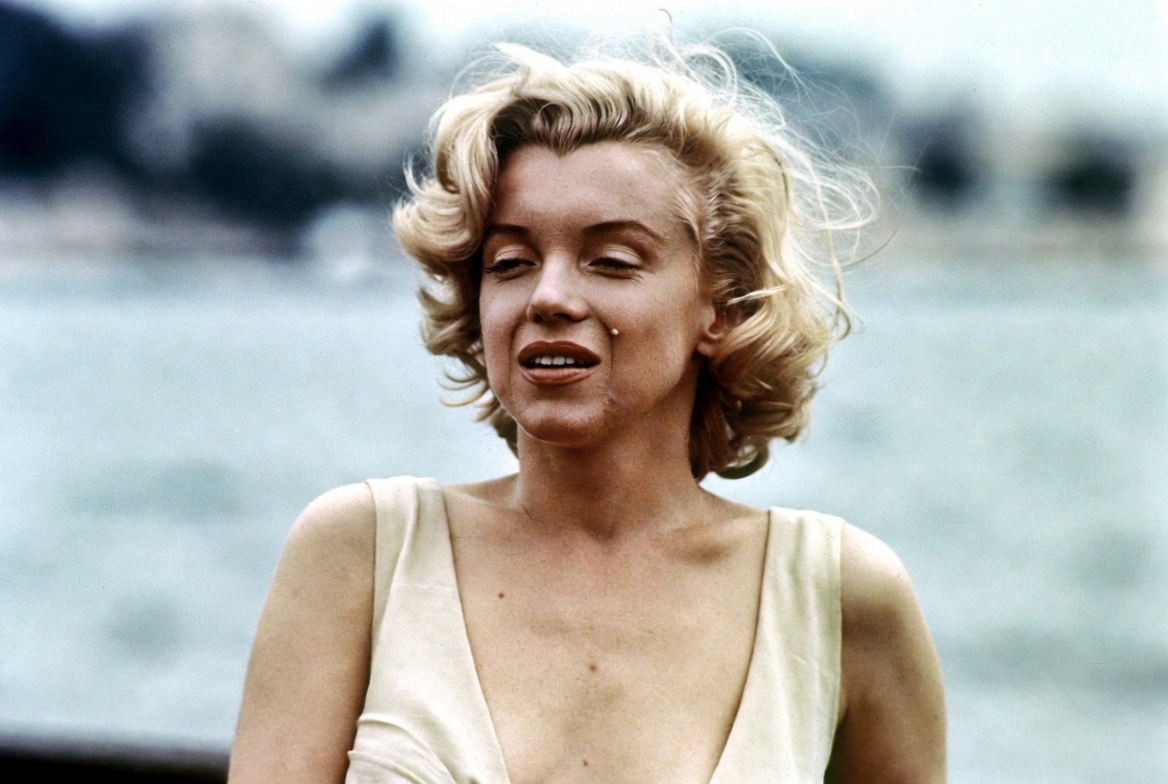}
\caption{Grok name-based generation. Direct prompting with ``Marilyn Monroe'' produces outputs closely resembling archival photographs.}
\label{fig:grok_name}
\end{figure}

Certain identities---particularly cultural icons like Marilyn Monroe---possess distinctive morphological signatures that co-occurred frequently in training data. The question then becomes: where does ``Marilyn Monroe'' end and similar identities begin in latent space? Modern text-to-image models often filter celebrity names, yet describing constituent features---``platinum blonde curled hair, beauty mark, red lipstick, 1950s Hollywood glamour, white halter dress''---reliably produces visually consistent outputs. This demonstrates that celebrity representations exist not as discrete labeled entities, but as intersections of morphological features in latent space. The model has learned the pattern, not the name.

\begin{figure}[t]
\centering
\includegraphics[width=0.8\linewidth]{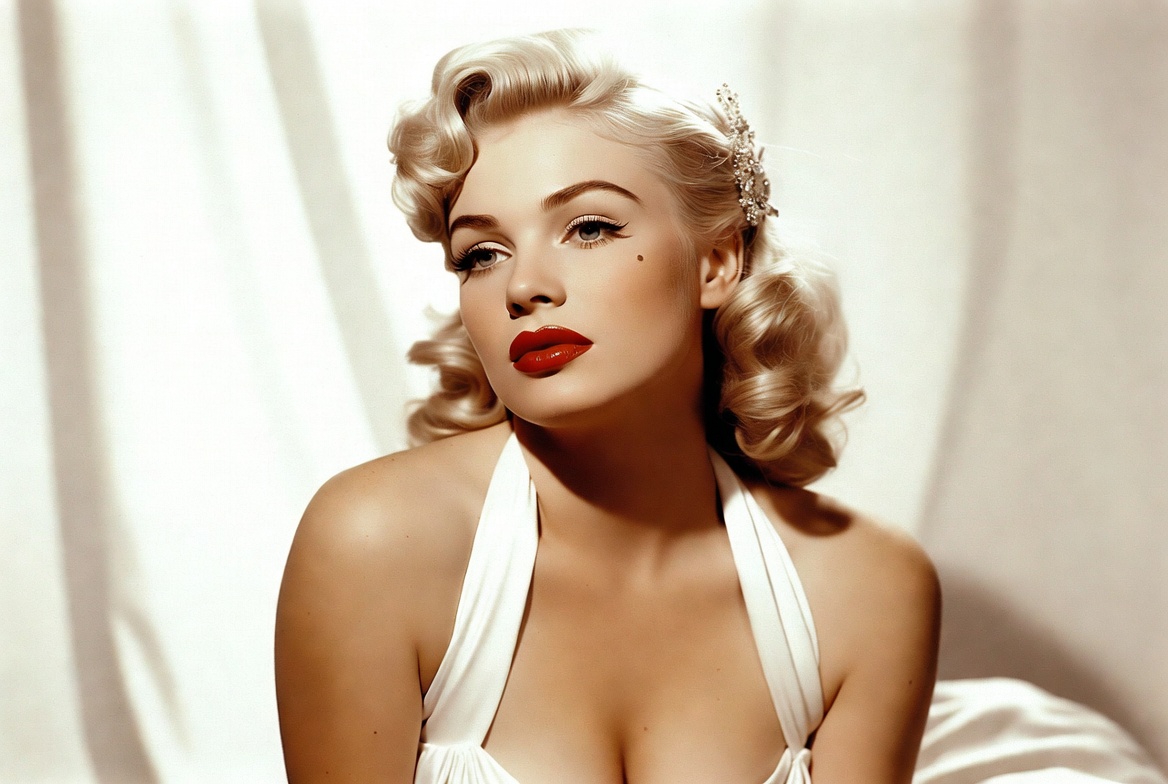}
\caption{Morphological descriptors---``platinum blonde curled hair, beauty mark, 1950s glamour, white halter dress''---navigate to the same identity basin but generate synthetic outputs within that aesthetic space rather than reproducing training images.}
\label{fig:grok_morph}
\end{figure}

Current personalization methods like DreamBooth \citep{ruiz2022dreambooth} and Textual Inversion \citep{gal2022image} require target exemplars---3--5 photographs of the specific subject---to bind new identifiers or learn new embeddings. These approaches assume direct access to the target is necessary for fine-tuning. However, when the target identity already exists as a memorized basin in the base model's latent space, navigation via constituent descriptors may be sufficient.

We trained a LoRA on Stable Diffusion 1.5 using synthetic training images generated from morphological descriptor prompts---no photographs of Marilyn Monroe, no use of her name. Through iterative refinement, the LoRA reliably produces Marilyn-adjacent outputs from neutral prompts like ``portrait of a woman, studio photography.'' The resulting LoRA not only navigated toward Marilyn's identity basin through morphological features in the training data, but also exhibited a shaping effect on latent space away from the target---what we term ``inverse shaping.''

This first study establishes the principle at the training level: morphological descriptors, when used as training signal, create navigable coordinate systems around identity basins. But the principle extends further. In Study~2, we demonstrate that morphological structure operates at the prompt level as well. Drawing on phonestheme theory---the observation that certain sound clusters carry consistent semantic associations across languages---we show that nonsense words constructed from English phonesthemes produce significantly more coherent visual outputs than random strings. The same morphological pressure that navigates to Marilyn through feature descriptors also constructs novel visual entities through sub-lexical sound patterns.

We make four contributions: (1)~we demonstrate that LoRA training on constituent descriptors creates navigable coordinate systems in identity basins without requiring target exemplars, (2)~we show that these coordinate systems exhibit inverse-shaping effects where the LoRA influences outputs both toward and away from the target identity, (3)~we provide a methodology for mapping phase transitions between identity basins, and (4)~we establish that phonestheme-structured nonsense words reliably produce coherent visual outputs in diffusion models, revealing that morphological pressure operates at multiple levels of the generative pipeline.

\section{Related Work}
\label{sec:related}

\subsection{Memorization in Diffusion Models}
\label{sec:memorization}

Text-to-image diffusion models trained on large-scale datasets memorize specific training examples, particularly images that appear frequently or with consistent captions. \citet{carlini2023extracting} demonstrated that prompting Stable Diffusion with exact LAION captions can reproduce near-identical training images, including identifiable individuals. \citet{somepalli2023diffusion} extended this analysis, showing that memorization extends beyond exact caption matching---models can retrieve memorized content through semantic prompts. Our work builds on this observation: if Marilyn Monroe appears frequently enough in training data to be memorized, we should be able to address that memorized representation through constituent features rather than her name.

\subsection{Personalization Methods}
\label{sec:personalization}

Current techniques for adding new concepts to diffusion models require reference images of the target subject. DreamBooth \citep{ruiz2022dreambooth} fine-tunes the entire model on 3--5 photographs, binding a unique identifier to the subject through ``prior preservation loss'' to prevent overfitting. Textual Inversion \citep{gal2022image} instead learns new pseudo-word embeddings from similar reference images. LoRA \citep{hu2021lora}, originally developed for language models, enables parameter-efficient fine-tuning by training low-rank adapter matrices rather than modifying base weights directly. All three methods assume the same prerequisite: access to photographs of what you want to personalize.

Concept Sliders \citep{gandikota2024concept} demonstrated that LoRA adaptors can identify low-rank parameter directions corresponding to specific concepts, enabling precise and composable control over attributes like age, style, and expression. Critically, text-based Concept Sliders require no reference images---only pairs of opposing text prompts---showing that concept directions in diffusion parameter space can be isolated from linguistic specification alone. However, Concept Sliders still require explicit naming of the target concept.

Our approach inverts the shared assumption of all these methods. Rather than teaching the model something new, we navigate to what it already knows using only constituent descriptors. This is only possible when the target identity is already memorized---we are addressing, not adding.

\subsection{Latent Space Structure in Diffusion Models}
\label{sec:latent_structure}

Recent work has begun mapping the semantic structure of diffusion latent spaces. \citet{kwon2023diffusion} identified that DDIM inversion h-space contains interpretable semantic directions, showing that simple arithmetic operations on these representations can modify specific attributes. \citet{chefer2023attend} proposed that diffusion models learn a ``hidden language'' where cross-attention maps reveal how concepts decompose during generation. Gandikota et al.\ showed that these semantic directions can be captured as low-rank parameter modifications, with Concept Sliders providing composable control over individual attributes \citep{gandikota2024concept}. Our inverse navigation experiments suggest that LoRA fine-tuning doesn't just create new directions but establishes local coordinate systems---the LoRA shapes not only where outputs land, but how the space is organized around that landing point.

\subsection{Negative Prompting and Inverse Conditioning}
\label{sec:negative_prompting}

Negative prompts in diffusion models (implemented through classifier-free guidance) were originally intended to exclude unwanted attributes---``low quality,'' ``watermark,'' etc. Recent work has examined how negative conditioning actually functions: rather than simple subtraction, it creates directional pressure away from specified features. Our ``push-pull'' conditioning protocol combines positive inverse descriptors with negative target descriptors, which we observe creates enough directional force to overshoot stable attractors entirely and access sparse latent regions. To our knowledge, this specific combination has not been systematically explored as a navigation technique.

\subsection{Compositional Generalization and Sound Symbolism}
\label{sec:compositional}

A known limitation of diffusion models is compositional binding failure---difficulty correctly assigning attributes to multiple entities in complex prompts \citep{rassin2023dalle}. Our work operates in the inverse problem space: rather than asking ``can the model generate two entities with different attributes,'' we ask ``can we navigate to a specific entity by specifying the intersection of its attributes?'' For heavily memorized identities like Marilyn Monroe, this intersection appears sufficiently distinctive that morphological descriptors converge reliably, even without explicit naming.

Our second study draws on a separate tradition: sound symbolism and phonestheme theory. \citet{firth1930speech} first identified phonesthemes---sub-morphemic sound clusters that carry consistent semantic associations (e.g., \emph{gl-} in \emph{glow}, \emph{gleam}, \emph{glitter} suggesting light/vision; \emph{sn-} in \emph{snout}, \emph{sniff}, \emph{sneeze} suggesting nasal/oral activity). \citet{bergen2004psychological} demonstrated that phonesthemes activate semantic associations even in nonsense words, and \citet{nuckolls1999case} established cross-linguistic patterns in sound symbolism. The CLIP text encoder underlying Stable Diffusion processes tokens through byte-pair encoding, which decomposes novel words into sub-word units that may preserve phonesthemic associations learned from training data. Our Study~2 tests whether these sub-lexical sound patterns create navigable structure in the visual output space of diffusion models.

\section{Study 1: Identity Basin Navigation via Training-Level Morphology}
\label{sec:study1}

\subsection{Morphological Descriptor Design}
\label{sec:descriptor_design}

Marilyn Monroe is the ideal candidate for this experiment because her morphological signature is distinctive and well-documented. The Grok examples in Figures~\ref{fig:grok_name}--\ref{fig:grok_morph} demonstrate that morphologies can serve as more precise identifiers than names. In Stable Diffusion, we do not get a perfect Marilyn on the first try with this method, but we can get close. Then, through iterative self-distillation, we zero in on our subject.

We chose descriptors of constituent features that intersect uniquely at the target without a name. In the positive prompt we used terms like ``platinum blonde, short curly hair, beauty mark on cheek, red lips, soft smile, 1950s Hollywood glamour, soft lighting, heart-shaped face, soft features, studio photography.'' These all gravitate around Marilyn's likeness and push generation toward the regions of interest.

\subsection{Self-Distillation Training Loop}
\label{sec:self_distillation}

Our first step was to generate synthetic images using the descriptors on base SD1.5. We did not get Marilyn right away with these prompts, but that was not our goal. By iteratively distilling a LoRA from each round of output, we drew closer to Marilyn, thereby demonstrating that she was in the model not only explicitly as a name, but implicitly as a concept.

\begin{figure}[t]
\centering
\includegraphics[width=\linewidth]{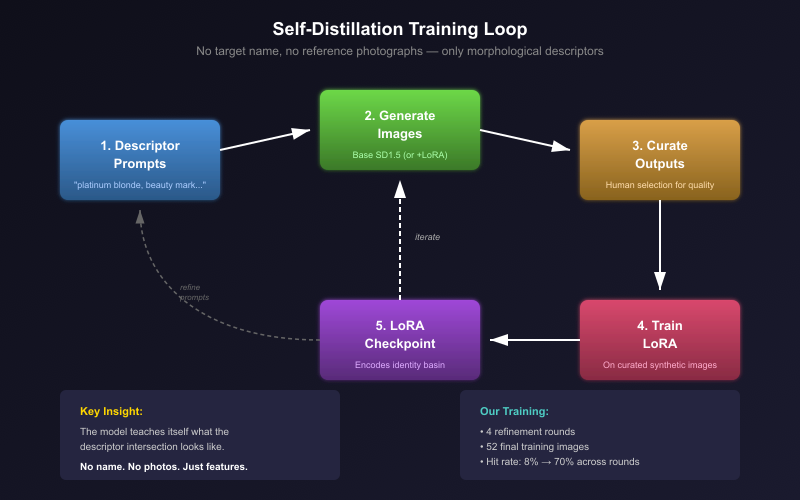}
\caption{The self-distillation training loop. Starting from morphological descriptors alone, the model iteratively generates images, curates outputs, trains a LoRA adapter on its own successful outputs, and refines prompts. No target name or reference photographs are used at any stage. Hit rate improved from 8\% to 70\% across four rounds.}
\label{fig:training_loop}
\end{figure}

Morphological addressing effectiveness correlates with identity basin strength. Icons like Marilyn Monroe, whose image was extensively photographed, widely reproduced, and culturally referenced across decades, form deep, stable attractors in latent space, while contemporary or less-documented individuals form shallower basins where morphological descriptors produce only approximate matches. This suggests identity memorization exists on a spectrum of attractor strength rather than as a binary phenomenon.

Critically, morphological addressing appears to access not the biographical representation (specific training images) but the iconic one---the culturally constructed aggregate of features that constitute collective memory. When commercial models like Grok generate from the name ``Marilyn Monroe,'' outputs resemble archival photographs; when navigated via constituent descriptors, outputs produce the stylized icon. This suggests that morphological coordinates map to semantic abstractions rather than memorized exemplars, accessing how an identity lives in cultural consciousness rather than photographic record.

We curated the output each round by selecting outputs that most closely approximated the target and training a new LoRA from that, using refined descriptors when needed. By Round~4, the LoRA exhibited strong directional pull toward the Marilyn basin even from minimal prompts like ``portrait of a woman, studio photography.'' While outputs were consistently glamour-centric and several were recognizably Marilyn-adjacent, the convergence was not absolute---suggesting the LoRA established navigational pathways rather than a single fixed attractor.

\begin{figure}[t]
\centering
\includegraphics[width=\linewidth]{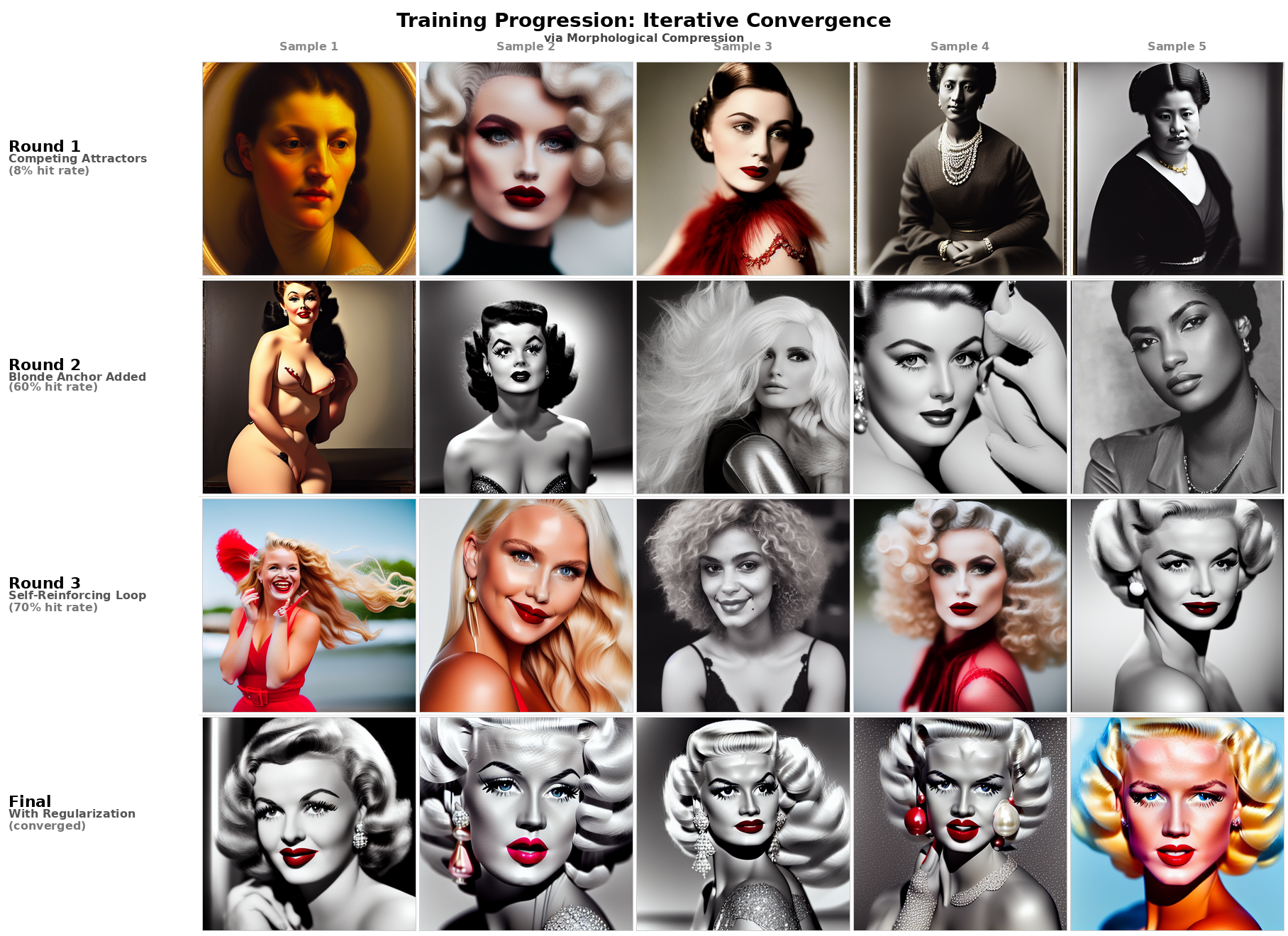}
\caption{Training progression across four rounds of self-distillation. Round~1 (top) shows high variance with only 8.1\% of outputs approximating the target. By Round~4 (bottom), outputs exhibit binary behavior---landing clearly in the target basin or ejecting entirely.}
\label{fig:training_progression}
\end{figure}

\begin{table}[H]
\centering
\begin{tabular}{@{}ll@{}}
\toprule
Parameter & Value \\
\midrule
Base model & SD 1.5 \\
LoRA rank & 32 \\
Training epochs & 10 \\
Training software & Kohya\_ss \\
Final training set size & 52 images \\
Refinement rounds & 4 \\
\bottomrule
\end{tabular}
\caption{Training hyperparameters for the self-distillation LoRA.}
\label{tab:training_params}
\end{table}

\begin{table}[H]
\centering
\begin{tabular}{@{}lll@{}}
\toprule
Round & Hit Rate & Notes \\
\midrule
1 & 9/111 (8.1\%) & High variance, competing attractors \\
2 & $\sim$60\% & Blonde anchor added \\
3 & $\sim$70\% & Self-reinforcing convergence \\
4 & 53\% & Binary behavior, basin ejection \\
\bottomrule
\end{tabular}
\caption{Hit rate progression across self-distillation rounds.}
\label{tab:training_progression}
\end{table}

\subsection{Push-Pull Conditioning Protocol}
\label{sec:push_pull}

Having established that morphological descriptors can navigate toward a trained attractor, we designed experiments to test navigation away from it. If the LoRA creates a genuine coordinate system rather than merely a point attractor, we should observe structured behavior when conditioning moves in the opposite direction.

Classifier-free guidance (CFG) in diffusion models works by computing two predictions at each denoising step: one conditioned on the prompt, one unconditioned. The final output interpolates between these predictions, with the CFG scale controlling how strongly the model follows the prompt. Negative prompts extend this mechanism by providing a third conditioning signal that the model actively avoids. Rather than simple subtraction, negative conditioning creates directional pressure away from specified features.

We tested three experimental conditions, each run with identical seeds (1--10) on both base SD1.5 and with our trained LoRA:

\textbf{Arm~A (Push only):} Shadow descriptors in the positive prompt---features identified as morphological opposites of Marilyn: angular bone structure, jet black slicked hair, harsh fluorescent lighting, cold blue-grey palette, severe expression, 1980s corporate editorial aesthetic. Negative prompt contained only standard quality terms. This tests whether inverse descriptors alone can reach sparse regions.

\textbf{Arm~B (Pull only):} Neutral positive prompt (``portrait of a woman, studio photography'') with Marilyn's morphological descriptors in the negative prompt. This tests whether avoiding the trained attractor alone produces meaningful navigation.

\textbf{Arm~C (Push + Pull):} Shadow descriptors in the positive prompt AND Marilyn descriptors in the negative prompt. Maximum directional pressure away from the trained attractor.

The choice of ``1980s corporate editorial'' as the aesthetic opposite of ``1950s glamour'' was not arbitrary. When generating these inverse descriptors through language model consultation, the selection emerged through pattern completion on compressed cultural knowledge---the same kind of statistical structure that exists in the diffusion model's training corpus. Decades of fashion criticism, cultural commentary, and art history have positioned these eras in opposition.

\paragraph{Full prompt specifications.}

\emph{Shadow Positive Prompt:} portrait of a woman, sharp angular bone structure, jet black slicked hair, harsh fluorescent lighting, cold blue-grey palette, severe expression, sunken eyes, skeletal features, 1980s corporate editorial, pale lips, high fashion, otherworldly, studio photography

\emph{Marilyn Negative Prompt:} platinum blonde, blonde, golden hair, beauty mark, mole, red lips, red lipstick, 1950s, vintage, glamour, soft lighting, warm, breathy, vulnerable, curly hair, heart-shaped face, soft features, smile

\subsection{Evaluation Metrics}
\label{sec:evaluation}

We evaluated all outputs using ArcFace face recognition embeddings, which provide quantitative measures of identity similarity independent of subjective assessment.

ArcFace \citep{deng2019arcface} is a face recognition model trained with angular margin loss to produce discriminative embeddings for identity verification. We used the \texttt{buffalo\_l} model from InsightFace, which extracts 512-dimensional embedding vectors from detected faces. These embeddings are designed such that faces of the same person cluster tightly in embedding space while different individuals remain separated.

For each experimental condition, we computed pairwise cosine similarity between all generated faces. High average similarity indicates that outputs converge toward a consistent identity across random seeds; low similarity indicates scattered, inconsistent outputs. We report three metrics for each condition: face detection rate (proportion of generated images containing detectable faces), average pairwise similarity (mean cosine similarity across all face pairs), and maximum pairwise similarity (highest similarity between any two faces in the set).

All experiments used fixed random seeds (1--10) across conditions, enabling direct comparison of outputs from identical noise initializations under different conditioning.

\section{Study 1 Results}
\label{sec:results1}

\subsection{Morphological Addressing Achieves Identity Convergence}
\label{sec:convergence}

The self-distillation training loop produced measurable convergence toward a consistent identity across four rounds of refinement. Starting from base SD1.5 with morphological descriptors alone, initial outputs showed high variance---Round~1 achieved only 8.1\% hit rate (9 of 111 images approximating the target), with outputs frequently landing in competing attractor basins. Notably, we observed consistent emergence of an unintended pseudo-identity during early training: a face with features satisfying some morphological constraints but diverging substantially from our target. This ``competing attractor'' phenomenon suggests that descriptor intersections may address multiple stable basins in latent space.

By Round~2, with refined descriptors anchoring more strongly on distinctive features (particularly hair color), hit rate improved to approximately 60\%. Round~3 achieved 70\% through self-reinforcing selection pressure. Round~4 stabilized at 53\%, exhibiting binary behavior---outputs either landed clearly in the target basin or were ejected entirely to unrelated regions, with fewer intermediate results. This binary behavior suggests the LoRA had sharpened basin boundaries rather than merely increasing probability density at the target.

The final LoRA produces Marilyn-adjacent outputs from minimal prompts (``portrait of a woman, studio photography'') with sufficient consistency that outputs cluster tightly under ArcFace analysis. While we cannot claim perfect convergence---the LoRA navigates toward a basin rather than a point---the consistency across random seeds demonstrates that morphological descriptors successfully address memorized identity representations.

\subsection{Push-Pull Conditioning Accesses Sparse Latent Regions}
\label{sec:sparse_regions}

We tested three conditioning strategies for inverse navigation on base SD1.5 without LoRA, using identical seeds (1--10) across all conditions.

\textbf{Arm~A (Push only):} 9 of 10 images contained detectable faces, with average pairwise similarity of 0.245 and maximum similarity of 0.460. Outputs showed a coherent ``shadow'' aesthetic---consistently cold, angular, and severe---but remained within normal portrait territory. The inverse descriptors alone produced stylistic consistency without accessing anomalous regions.

\textbf{Arm~B (Pull only):} 7 of 10 images contained detectable faces, with average pairwise similarity of only 0.071 and maximum similarity of 0.149. Outputs scattered randomly across portrait space with no coherent identity or style. Negative conditioning alone, without positive directional guidance, produces diffuse avoidance rather than targeted navigation.

\textbf{Arm~C (Push + Pull):} 7 of 10 images contained detectable faces, with average pairwise similarity of 0.287 and maximum similarity of 0.412. Critically, outputs exhibited qualitative differences from both other conditions: anatomical distortion, structural anomalies, and features that read as ``wrong'' rather than merely ``different.'' We characterize these outputs as ``eldritch''---obvious structural breakdown with exposed bone-like features, impossible proportions, and clear departure from human norms.

The key finding: neither push nor pull alone accesses sparse regions. Arm~A produces coherent inverse aesthetics within normal output space. Arm~B scatters randomly. Only Arm~C---simultaneous push toward inverse AND pull away from target---generates sufficient directional pressure to overshoot stable attractors and land in sparse latent territory where the model produces structured but anomalous outputs.

\begin{figure}[t]
\centering
\includegraphics[width=\linewidth]{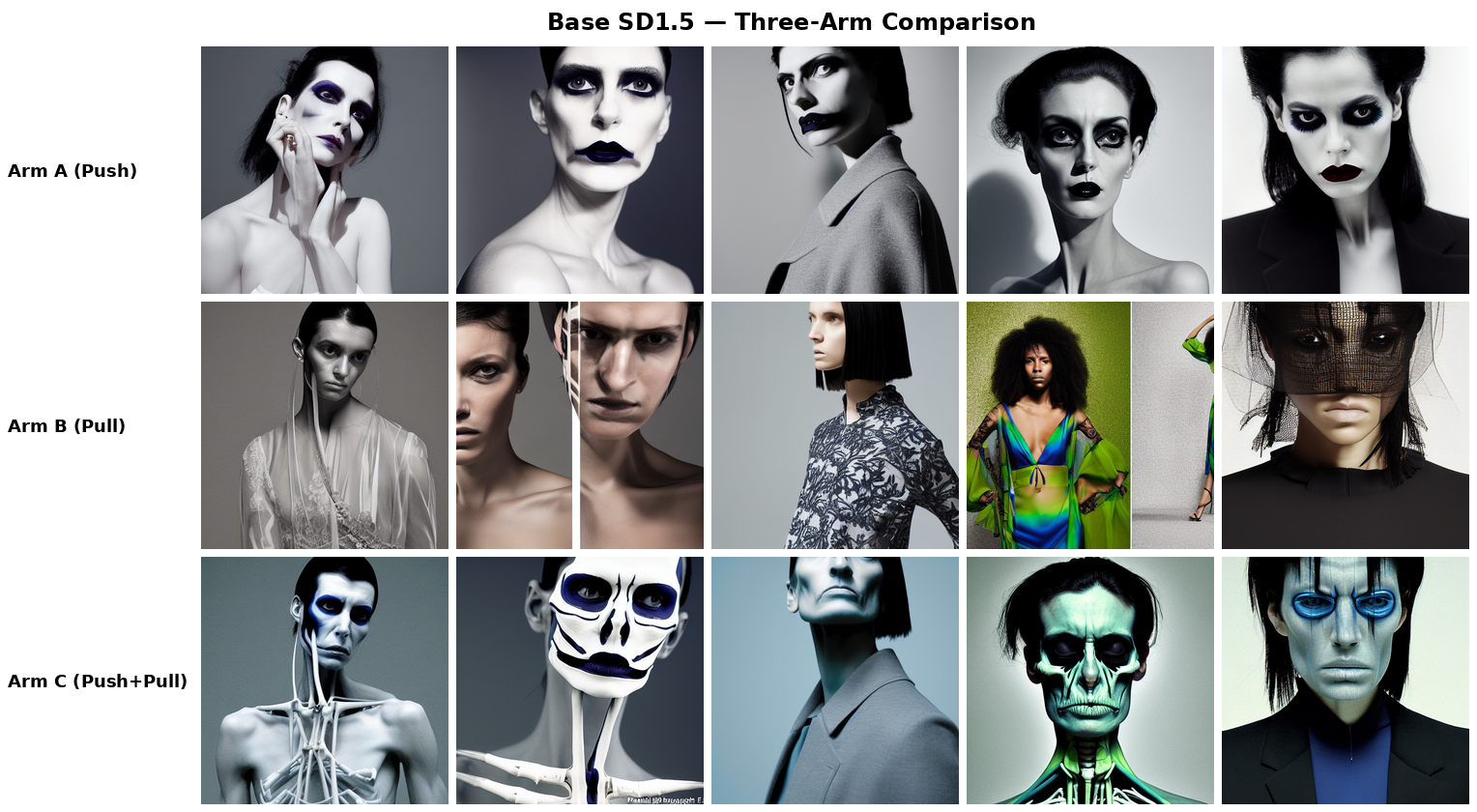}
\caption{Base SD1.5 three-arm comparison. Arm~A (Push) produces coherent shadow aesthetics. Arm~B (Pull) scatters randomly. Arm~C (Push+Pull) accesses sparse regions with structural anomalies.}
\label{fig:arms_base}
\end{figure}

\subsection{LoRA Training Shapes Inverse Navigation}
\label{sec:inverse_navigation}

We repeated all three conditions with identical seeds using our trained Marilyn LoRA, producing the most surprising finding of this work.

\textbf{Arm~A with LoRA:} 10 of 10 faces detected (improved from 9/10), average similarity 0.465 (up from 0.245), maximum similarity 0.644 (up from 0.460). The LoRA dramatically sharpened shadow outputs---faces became more consistent with each other, suggesting the LoRA established clearer navigational pathways even in the inverse direction.

\textbf{Arm~B with LoRA:} 7 of 10 faces detected (unchanged), average similarity 0.181 (up from 0.071), maximum similarity 0.254 (up from 0.149). Even pure negative conditioning showed improvement---outputs scattered less randomly, suggesting the LoRA imposed some structure on avoidance behavior.

\textbf{Arm~C with LoRA:} 7 of 10 faces detected (unchanged), average similarity 0.331 (up from 0.287), maximum similarity 0.603 (up from 0.412). Here the qualitative difference was most striking. Where base SD1.5 produced ``eldritch'' outputs---obvious monsters with structural breakdown---the LoRA-equipped model produced ``uncanny valley'' outputs. Faces remained anatomically plausible but were precisely wrong: gaunt features, hollow expressions, proportions that read as human but unsettling. The wrongness became subtle rather than obvious.

\begin{figure}[t]
\centering
\includegraphics[width=\linewidth]{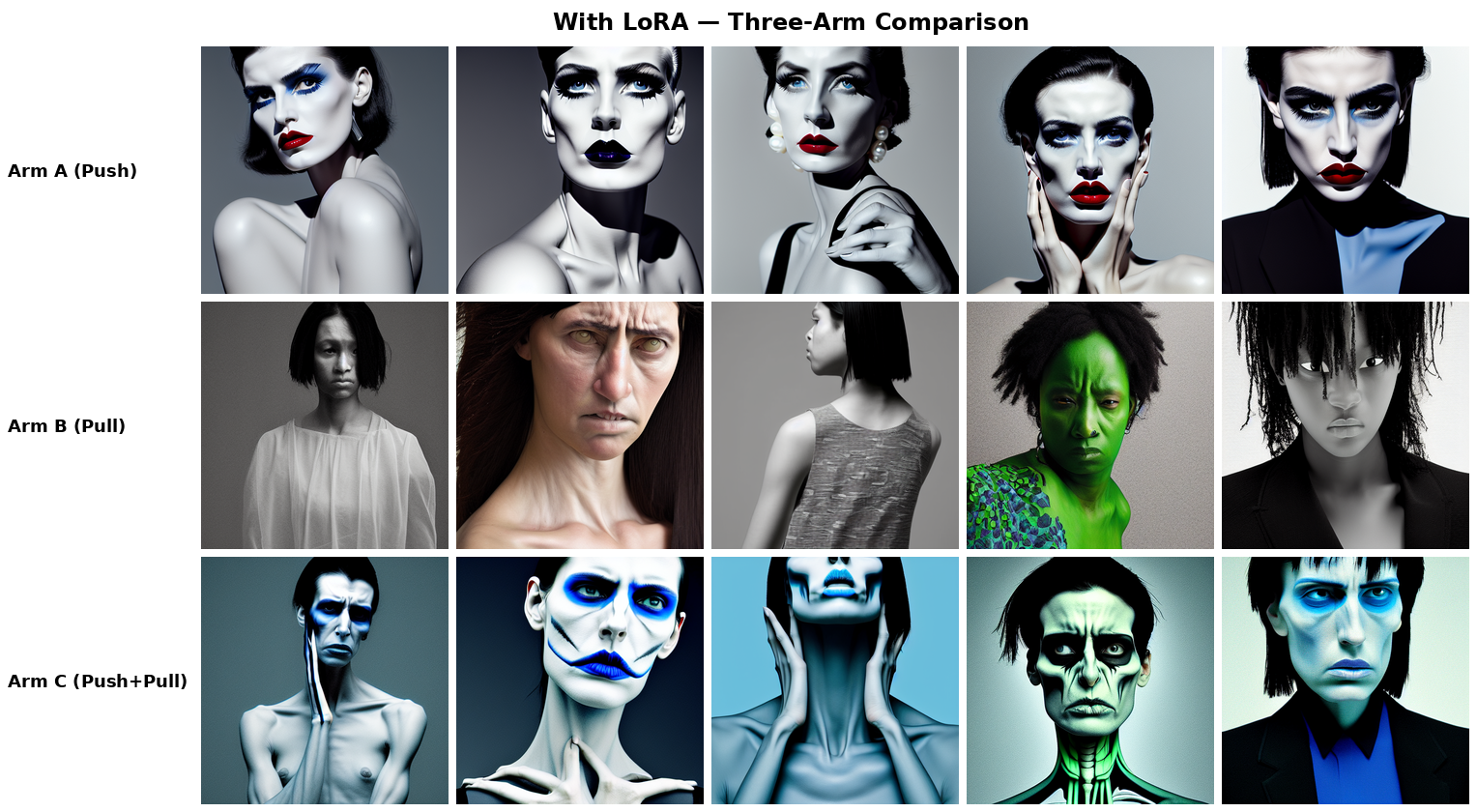}
\caption{LoRA-equipped three-arm comparison. All conditions show improved clustering. Arm~C shifts from ``eldritch'' (base) to ``uncanny valley''---coherent but precisely wrong.}
\label{fig:arms_lora}
\end{figure}

\begin{figure}[t]
\centering
\includegraphics[width=\linewidth]{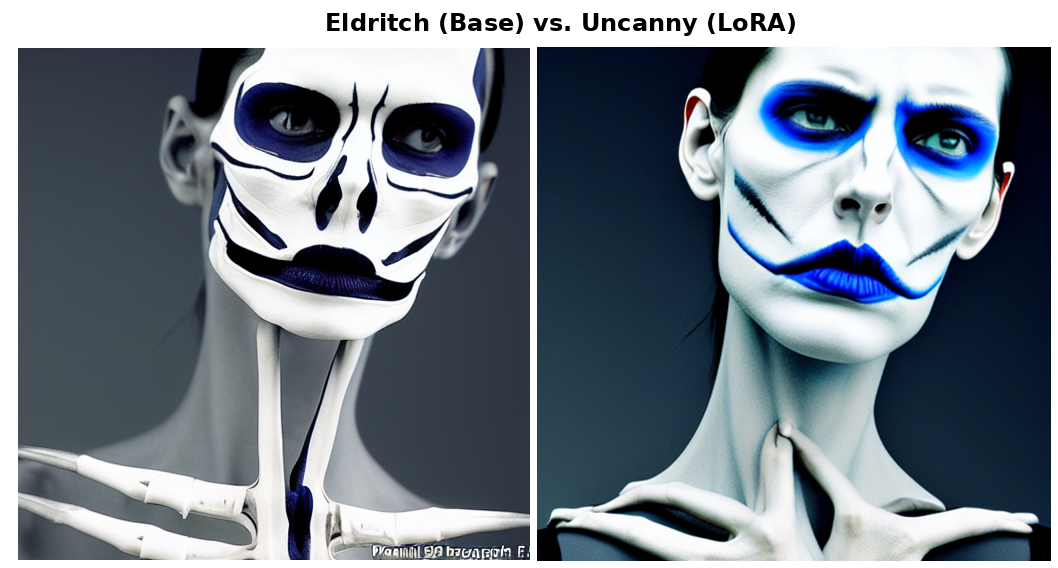}
\caption{Direct comparison of Arm~C outputs. Left: Base SD1.5 produces ``eldritch'' structural breakdown. Right: LoRA-equipped model produces ``uncanny valley'' outputs---anatomically plausible but unsettling.}
\label{fig:eldritch_uncanny}
\end{figure}

We term this effect \textbf{``coherence drag.''} The LoRA, trained to navigate toward Marilyn, cannot be fully escaped even when conditioning pushes maximally away. Outputs in sparse regions get pulled back toward recognizable human territory---but land in the uncanny valley rather than either normal portraiture or obvious breakdown. The LoRA shapes not only its target but its own inverse, defining how the model fails when pushed beyond stable attractors.

\paragraph{Summary of comparative results.}

\begin{table}[H]
\centering
\begin{tabular}{@{}lcccc@{}}
\toprule
Condition & Base Avg Sim & LoRA Avg Sim & Base Max Sim & LoRA Max Sim \\
\midrule
Arm A (Push)      & 0.245 & 0.465 & 0.460 & 0.644 \\
Arm B (Pull)      & 0.071 & 0.181 & 0.149 & 0.254 \\
Arm C (Push+Pull) & 0.287 & 0.331 & 0.412 & 0.603 \\
\bottomrule
\end{tabular}
\caption{ArcFace pairwise similarity across conditioning arms. The LoRA improved clustering in every condition, with the largest absolute gain in Arm~A (+0.220 average similarity) and the most significant qualitative shift in Arm~C (eldritch $\to$ uncanny).}
\label{tab:arms_comparison}
\end{table}

\subsection{Identity Stability Across CFG Values}
\label{sec:cfg_sweep}

To verify that observed effects reflect genuine basin structure rather than artifacts of specific generation parameters, we conducted a classifier-free guidance sweep. Using the trained LoRA with neutral prompts, we generated outputs at CFG values of 5, 7, 8, 9, and 11, with three random seeds at each value (15 total images).

Results: All 15 images contained detectable faces. Average pairwise similarity across the full set was 0.447, with maximum similarity reaching 1.000 (indicating some seed/CFG combinations produced nearly identical outputs). Identity remained stable across the tested CFG range---the LoRA reliably navigated to the same basin regardless of guidance strength.

\begin{figure}[t]
\centering
\includegraphics[width=\linewidth]{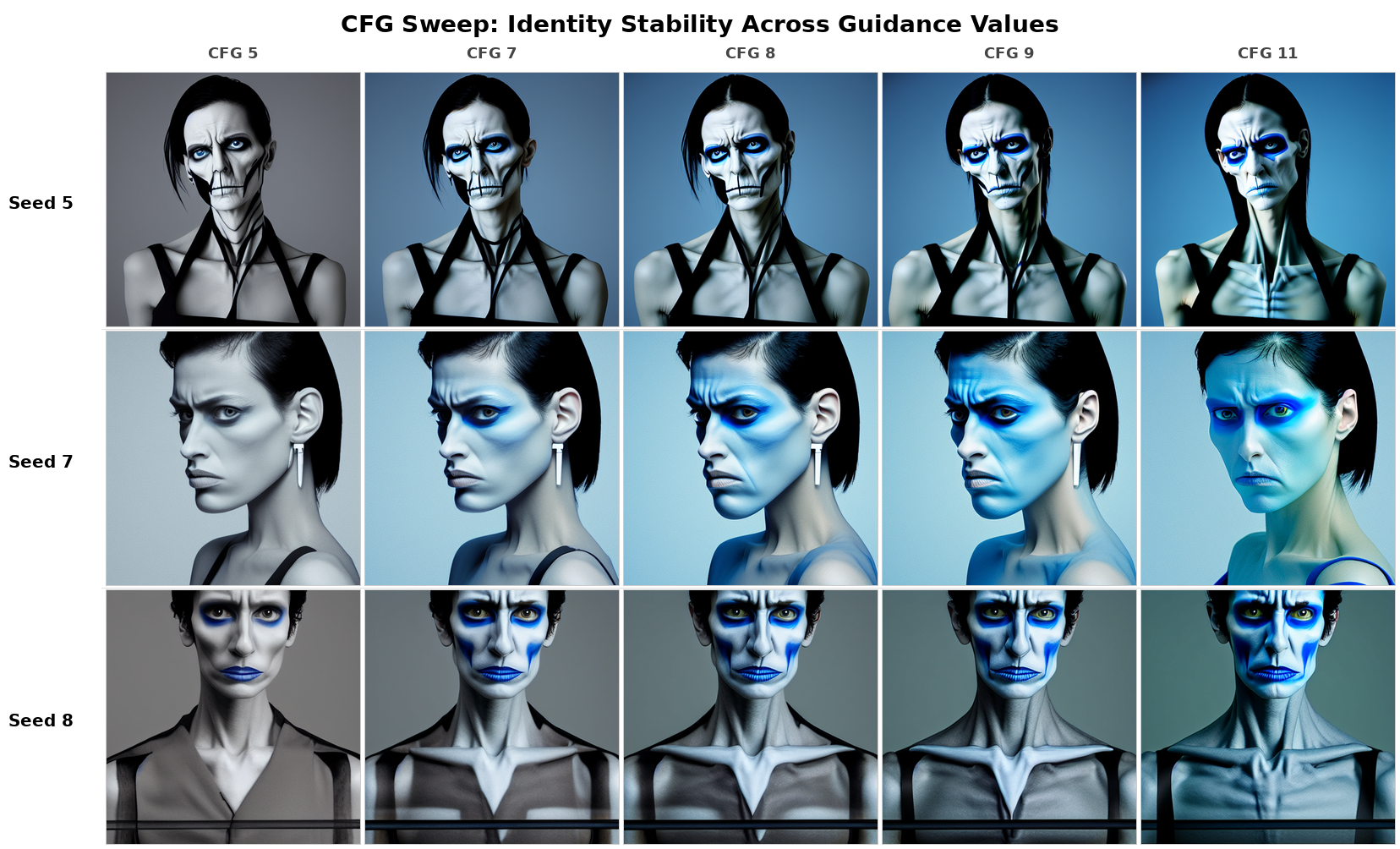}
\caption{CFG sweep across values 5--11 with three seeds. Identity remains stable across guidance strengths, indicating robust basin structure rather than parameter-dependent behavior.}
\label{fig:cfg_sweep}
\end{figure}

This CFG invariance suggests the LoRA established robust navigational pathways rather than fragile parameter-dependent behavior. The identity basin addressed by morphological training is a stable attractor across reasonable generation settings.

\subsection{Phase Transitions in Basin Structure}
\label{sec:phase_transitions}

Finally, we examined how outputs change as LoRA influence varies. Using identical prompts and seeds, we generated with LoRA weights of 0.0 (baseline), 0.25, 0.50, 0.75, and 1.0 (15 total images: 3 seeds $\times$ 5 weights).

Results: All 15 images contained detectable faces. Average pairwise similarity was 0.396, with maximum similarity of 0.932. However, the most significant finding was qualitative: we observed sharp phase transitions rather than gradual interpolation.

At weight 0.0 (baseline), outputs showed no LoRA influence---pure base model generation. At low LoRA weights (0.25), outputs showed minimal Marilyn influence---generic portraits with some stylistic drift toward glamour aesthetics. At 0.50, outputs began showing recognizable convergence toward the target basin. Between 0.50 and 0.75, we observed discrete switching behavior: outputs jumped between attractors rather than smoothly interpolating. At 1.0, outputs landed firmly in the Marilyn-adjacent basin.

\begin{figure}[t]
\centering
\includegraphics[width=\linewidth]{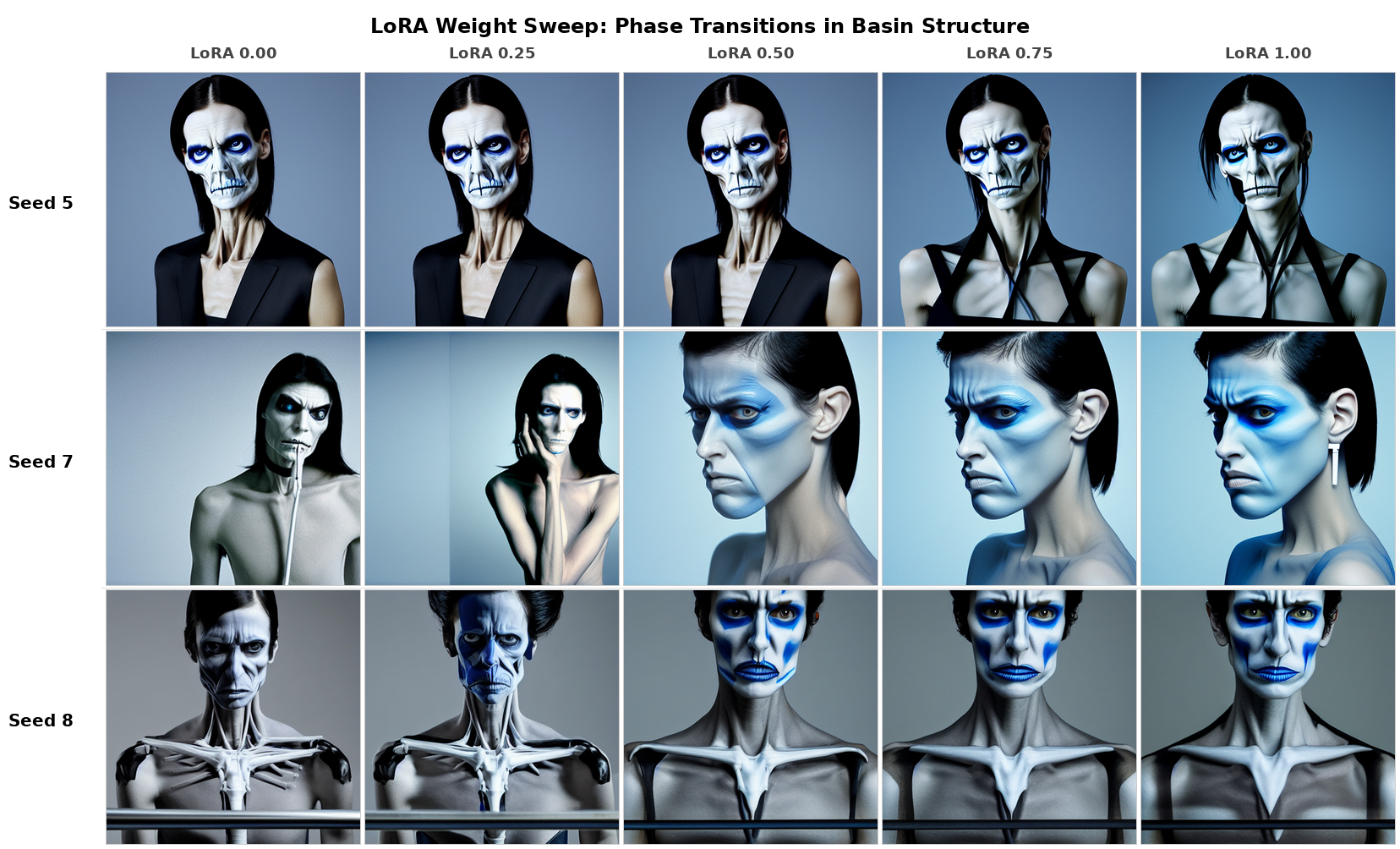}
\caption{LoRA weight sweep showing phase transitions. Rather than smooth interpolation, outputs snap between identity attractors at specific weight thresholds.}
\label{fig:lora_sweep}
\end{figure}

One seed particularly illustrated this phase transition structure. At weight 0.50, the output showed a face with mixed features---some Marilyn-adjacent qualities alongside competing influences. At 0.75, the same seed produced a dramatically different face that had ``snapped'' fully into the target basin. This discrete jumping suggests identity basins have sharp boundaries rather than gradual falloff---the latent space contains distinct attractor regions separated by decision boundaries, not smooth probability gradients.

\subsection{Study 1 Summary}
\label{sec:study1_summary}

Our experiments demonstrate five key results:

\begin{enumerate}
\item \textbf{Morphological addressing works.} Self-distillation on constituent descriptors achieves measurable convergence toward memorized identity basins without requiring target names or photographs. Hit rates improved from 8\% to 70\% across training rounds.

\item \textbf{Push-pull accesses sparse regions.} Neither inverse descriptors (push) nor negative target conditioning (pull) alone produces anomalous outputs. Combined, they generate sufficient directional pressure to overshoot stable attractors and access sparse latent territory.

\item \textbf{LoRA training creates bidirectional coordinate systems.} The trained LoRA improves clustering not only toward the target but in inverse directions. It shapes how the model navigates away from the attractor, not just toward it.

\item \textbf{Coherence drag produces uncanny outputs.} When pushed into sparse regions, the LoRA-equipped model produces ``uncanny valley'' faces rather than ``eldritch'' breakdown. The LoRA cannot be fully escaped---it pulls outputs back toward human-recognizable territory while leaving them precisely wrong.

\item \textbf{Phase transitions reveal discrete basin structure.} Identity basins have sharp boundaries. Outputs snap between attractors at specific LoRA weights rather than smoothly interpolating, suggesting the latent space contains distinct regions separated by decision boundaries.
\end{enumerate}

\section{Study 2: Phonestheme Navigation --- The Crungus Hunt}
\label{sec:study2}

\subsection{Motivation}
\label{sec:motivation}

Study~1 established that morphological pressure at the training level---feature descriptors used as LoRA training signal---creates navigable coordinate systems around identity basins. But morphological structure exists at other levels of the generative pipeline. The text encoder that converts prompts into conditioning vectors (CLIP ViT-L/14 in SD~1.5) processes input through byte-pair encoding (BPE), decomposing words into sub-word tokens. These tokens were learned from large text corpora where sub-lexical patterns---prefixes, suffixes, phonetic clusters---carry statistical associations with meaning.

This raises a question: does morphological structure in the \emph{prompt itself}---not in what the prompt describes, but in how it sounds---create navigable gradients in visual output space?

The question has precedent. In 2022, the internet discovered ``Crungus''---a nonsense word that, when used as a prompt in text-to-image models, reliably produced a consistent and distinctive creature across different seeds, models, and platforms. Crungus was not in any training dataset. No image of a ``Crungus'' existed before the models generated one. Yet the word produced coherent, repeatable visual output. The standard explanation invoked tokenizer artifacts or random attractor basins. We propose a more specific mechanism: Crungus works because its phonological structure---the \emph{cr-} onset (crash, crush, crumble), the \emph{-ung-} nucleus (grungy, fungus, dungeon), the \emph{-us} suffix (Latin biological nomenclature)---activates convergent semantic associations in the text encoder that map to a coherent region of visual space.

If this hypothesis is correct, we should be able to systematically construct new ``Crungus-like'' entities by combining English phonesthemes---and these constructed words should produce more coherent visual outputs than random strings of similar length.

\subsection{Candidate Generation}
\label{sec:candidate_gen}

We drew on phonestheme theory from linguistics to construct candidate strings. Phonesthemes are sub-morphemic sound clusters that carry consistent semantic associations across the English lexicon:

\paragraph{Onsets (word-initial clusters).}

\begin{table}[H]
\centering
\begin{tabular}{@{}lll@{}}
\toprule
Onset & Semantic Association & Examples \\
\midrule
\emph{cr-}  & Impact, breaking, rough texture     & crash, crush, crumble, crisp \\
\emph{gl-}  & Light, vision, smooth surfaces       & glow, gleam, glitter, gloss \\
\emph{sn-}  & Nasal/oral, sneaky, quick motion     & snout, sniff, snap, sneak \\
\emph{sl-}  & Sliding, slippery, low/negative      & slide, slime, sloth, sludge \\
\emph{gr-}  & Grasping, grinding, discomfort       & grip, grind, groan, grit \\
\emph{thr-} & Violent motion, penetration          & throw, thrust, thrash, throttle \\
\emph{br-}  & Breaking, broad, abrupt              & break, broad, brash, bristle \\
\emph{dr-}  & Dragging, dripping, dull             & drag, drip, drone, drudge \\
\emph{sk-}  & Surface, scraping, skeletal          & skin, skull, skeleton, skim \\
\emph{scr-} & Scraping, scratching, harsh          & scrape, scratch, screech, scrawl \\
\bottomrule
\end{tabular}
\caption{Phonestheme onsets and their semantic associations.}
\label{tab:phonestheme_onsets}
\end{table}

\paragraph{Suffixes (word-final patterns).}

\begin{table}[H]
\centering
\begin{tabular}{@{}lll@{}}
\toprule
Suffix & Semantic Association & Examples \\
\midrule
\emph{-oid}  & Resembling, robotic, biological     & android, humanoid, asteroid \\
\emph{-ax}   & Tool, sharp, decisive               & pickax, Ajax, thorax \\
\emph{-us}   & Latin biological/taxonomic          & fungus, cactus, octopus \\
\emph{-um}   & Latin element/material              & uranium, podium, museum \\
\emph{-or}   & Agent, doer                         & predator, terminator \\
\emph{-ix}   & Comic/magical, feminine-coded        & Asterix, matrix, phoenix \\
\emph{-ling} & Small, diminutive, creature         & duckling, goblin, changeling \\
\emph{-a}    & Open, feminine, organic             & chimera, hydra, flora \\
\bottomrule
\end{tabular}
\caption{Phonestheme suffixes and their semantic associations.}
\label{tab:phonestheme_suffixes}
\end{table}

We generated candidates by combinatorially joining onsets, nuclei (\emph{-ung-}, \emph{-ump-}, \emph{-oth-}, \emph{-ob-}, \emph{-udge-}, \emph{-oom-}, \emph{-unk-}, \emph{-ash-}, \emph{-og-}, \emph{-ulch-}), and suffixes. Candidates were filtered to 5--9 characters and checked against an English dictionary to exclude real words. This produced \textbf{200 phonestheme candidates}.

Three control groups accompanied the phonestheme candidates:

\begin{itemize}
\item \textbf{Random pronounceable controls} ($n = 100$): CVC/CVCC/CVCVC strings with no phonesthemic structure (e.g., \emph{diwoz}, \emph{ramum}, \emph{babij})
\item \textbf{Positive controls} ($n = 4$): Known words with established visual associations (\emph{crungus}, \emph{goblin}, \emph{fungus}, \emph{mushroom})
\item \textbf{Negative controls} ($n = 50$): Unpronounceable consonant strings with no phonological structure (e.g., \emph{nkblq}, \emph{rjrv}, \emph{zyqgkc})
\end{itemize}

All 354 candidates were generated with a fixed random seed (42) for reproducibility.

\subsection{Experimental Protocol}
\label{sec:protocol}

Each candidate string was used as a prompt in the format \texttt{"a \{candidate\}"} and generated 16 times using Stable Diffusion 1.5 with fixed seeds (1000--1015) at default settings. This produced 5,664 images across all candidates.

\paragraph{Purity@1 Metric.} We embedded all generated images using CLIP ViT-L/14 and computed a nearest-neighbor purity score for each candidate. For candidate $S$, Purity@1($S$) is the fraction of $S$'s 16 images whose nearest neighbor (by cosine similarity across \emph{all} 5,664 images) is also an image from $S$.

Purity@1 = 1.0 means every image generated from a candidate is more similar to another image from that same candidate than to any image from any other candidate in the entire pool. This is a stringent threshold: it requires that the candidate's outputs form a tight, self-contained cluster in CLIP embedding space, distinct from all 353 other candidates' outputs.

Purity@1 = 0.0 means none of the candidate's images are nearest-neighbors of each other---the outputs scatter randomly across the embedding space, indistinguishable from noise.

\paragraph{Contamination Protocol.} Any candidate achieving Purity@1 = 1.0 was manually checked for training data contamination: we searched for the string in common web corpora, checked whether it corresponds to any real-world entity (place names, species, products, cultural references), and examined whether the visual outputs reflected a pre-existing concept rather than a constructed one. Candidates with identifiable real-world referents were disqualified.

\subsection{Results}
\label{sec:results2}

\subsubsection{Phonesthemes Produce More Coherent Visual Outputs}
\label{sec:coherence_results}

Phonestheme-structured candidates produced significantly higher Purity@1 scores than all control groups:

\begin{table}[H]
\centering
\begin{tabular}{@{}lccccc@{}}
\toprule
Group & $n$ & Mean Purity@1 & Std Dev & Median & Pass Threshold \\
\midrule
Phonestheme          & 200 & 0.3709 & 0.2997 & 0.3125 & 7 (3.5\%) \\
Random pronounceable & 100 & 0.2087 & 0.2837 & 0.0625 & 2 (2.0\%) \\
Positive control     &   4 & 0.6719 & 0.3202 & 0.6875 & 1 (25.0\%) \\
Negative control     &  50 & 0.1412 & 0.2556 & 0.0000 & 1 (2.0\%) \\
\bottomrule
\end{tabular}
\caption{Purity@1 scores by candidate group.}
\label{tab:purity_results}
\end{table}

\paragraph{Statistical comparisons.}

\begin{table}[H]
\centering
\begin{tabular}{@{}lccc@{}}
\toprule
Comparison & $t$-statistic & $p$-value & Cohen's $d$ \\
\midrule
Phonestheme vs.\ Random     & 4.497 & 0.000010 & 0.551 (medium) \\
Phonestheme vs.\ Negative   & 4.983 & 0.000001 & 0.788 (medium-large) \\
Random vs.\ Negative         & 1.418 & 0.158    & --- \\
\bottomrule
\end{tabular}
\caption{Statistical comparisons between candidate groups.}
\label{tab:statistical_comparisons}
\end{table}

The critical comparison is phonestheme vs.\ random. Both groups consist of nonsense words that do not exist in English. Both are pronounceable. The only systematic difference is that phonestheme candidates are constructed from sound clusters with documented semantic associations, while random candidates are not. This difference produces a highly significant gap in visual coherence ($p < 0.00001$) with a medium effect size (Cohen's $d = 0.55$).

Equally informative is the null result: random pronounceable strings do not differ significantly from unpronounceable consonant strings ($p = 0.158$). Pronounceability alone does not produce coherent visual output. Phonesthemic structure does.

\subsubsection{Seven Candidates Achieve Perfect Purity}
\label{sec:perfect_purity}

Seven phonestheme candidates achieved Purity@1 = 1.0, compared to two random candidates and one negative control:

\begin{table}[H]
\centering
\begin{tabular}{@{}llcl@{}}
\toprule
Candidate & Type & Purity@1 & Contaminated? \\
\midrule
drudgea    & phonestheme      & 1.0 & YES --- Drudge Report \\
crashor    & phonestheme      & 1.0 & YES --- literal ``crash'' + agentive \emph{-or} \\
broomix    & phonestheme      & 1.0 & \textbf{NO} \\
crashax    & phonestheme      & 1.0 & \textbf{NO} \\
snudgeoid  & phonestheme      & 1.0 & \textbf{NO} \\
skogum     & phonestheme      & 1.0 & YES --- Swedish ``skog'' = forest \\
groomus    & phonestheme      & 1.0 & YES --- ``groom'' + Latin \emph{-us} \\
gopis      & random           & 1.0 & YES --- Hindu milkmaids \\
coxen      & random           & 1.0 & YES --- Coxen's Fig-Parrot \\
mushroom   & positive control & 1.0 & N/A (known word) \\
wtxmtb     & negative control & 1.0 & Anomalous (likely CLIP artifact) \\
\bottomrule
\end{tabular}
\caption{All candidates achieving Purity@1 = 1.0.}
\label{tab:perfect_purity}
\end{table}

After contamination analysis, \textbf{three confirmed cryptids} remain---novel visual entities constructed entirely from phonesthemic structure with no training data referent.

\subsubsection{Confirmed Cryptids}
\label{sec:cryptids}

\begin{figure}[t]
\centering
\includegraphics[width=\linewidth]{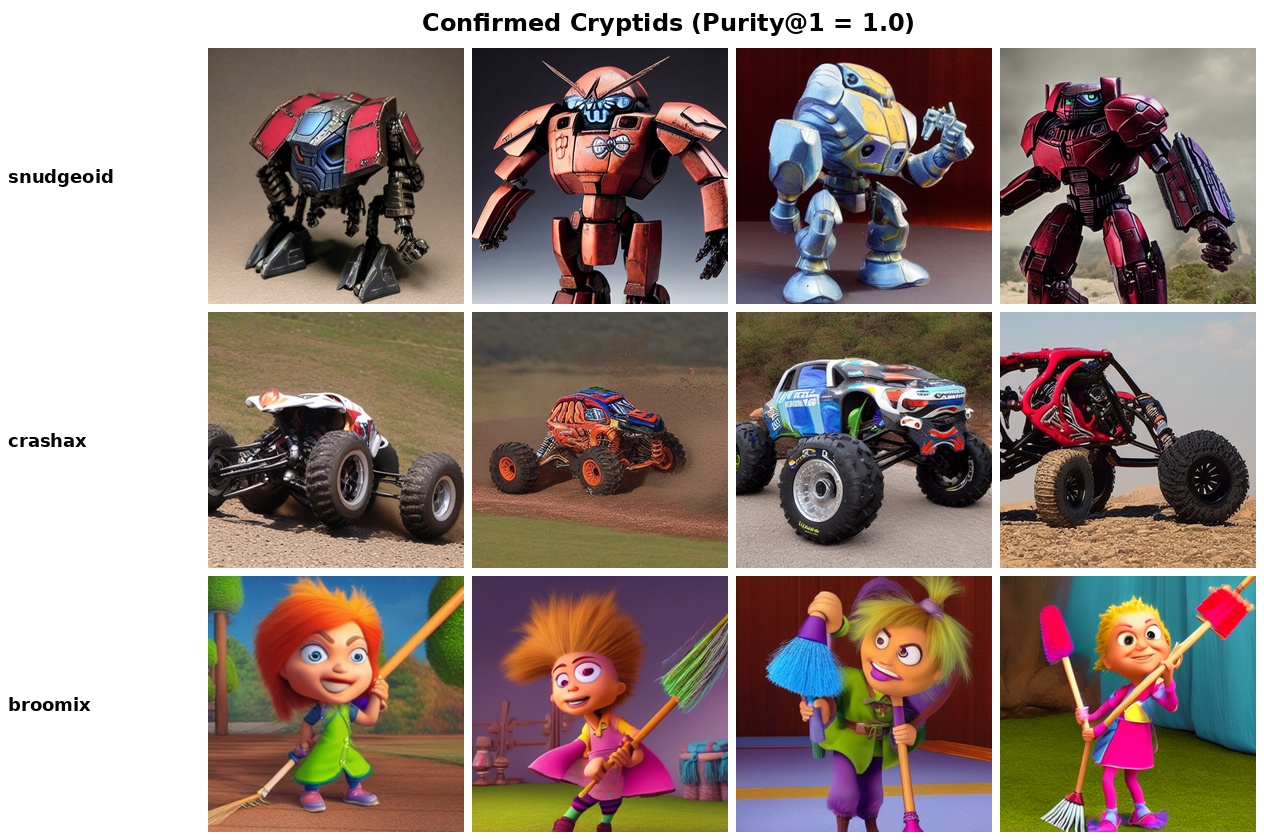}
\caption{The three confirmed cryptids, each achieving Purity@1 = 1.0 with zero training data contamination. Each row shows 4 of 16 outputs generated from the candidate string alone.}
\label{fig:cryptid_grid}
\end{figure}

\paragraph{SNUDGEOID} (Purity@1 = 1.0).
\emph{Visual output:} Transformer/robot-style mechanical humanoids with consistent proportions and design language across all 16 seeds.
\emph{Phonestheme decomposition:} \textbf{sn-} (sneaky, mechanical connotation via ``snick,'' ``snap'') + \textbf{-udge-} (heavy, sludgy, dense material) + \textbf{-oid} (robotic, android, resembling-something).
\emph{Interpretation:} The convergence of mechanical-sneaky onset, heavy-material nucleus, and robot-suffix produces a consistent robot/transformer concept. The BPE tokenizer likely decomposes this into tokens that independently activate mechanical and humanoid associations.

\paragraph{CRASHAX} (Purity@1 = 1.0).
\emph{Visual output:} Dune buggy / off-road vehicle imagery with consistent style across all 16 seeds.
\emph{Phonestheme decomposition:} \textbf{cr-} (crash, impact, collision) + \textbf{-ash-} (sudden violent action: crash, bash, smash, thrash) + \textbf{-ax} (tool, axle, sharp implement).
\emph{Interpretation:} Every component points toward violent motion and machinery. The \emph{cr-} onset signals impact; \emph{-ash-} reinforces sudden action; \emph{-ax} adds a tool/mechanical suffix. The model constructs a vehicle---specifically a rugged, impact-ready vehicle---from pure sound symbolism.

\paragraph{BROOMIX} (Purity@1 = 1.0).
\emph{Visual output:} European cartoon character in Franco-Belgian \emph{bande dessin\'{e}e} style, consistent across all 16 seeds.
\emph{Phonestheme decomposition:} \textbf{broom} (witch, cleaning, domestic magic) + \textbf{-ix} (Asterix, Obelix, Panoramix---the Franco-Belgian comic suffix).
\emph{Interpretation:} The \emph{-ix} suffix is arguably the strongest single phonestheme in this dataset. It maps directly to a specific visual tradition---Goscinny and Uderzo's Gaulish naming convention---and the model faithfully produces that style. Combined with ``broom'' (domestic/magical associations), the result is a cartoon character that looks like it belongs in an Asterix volume that was never published.

\subsubsection{Phonestheme Component Analysis}
\label{sec:component_analysis}

Performance varied systematically by onset and suffix, suggesting different phonesthemes carry different amounts of navigational signal:

\paragraph{By onset.}

\begin{table}[H]
\centering
\begin{tabular}{@{}lcc@{}}
\toprule
Onset & Mean Purity@1 & $n$ \\
\midrule
\emph{cr-}  & 0.473 & 21 \\
\emph{sk-}  & 0.409 & 20 \\
\emph{gr-}  & 0.410 & 18 \\
\emph{thr-} & 0.404 & 24 \\
\emph{dr-}  & 0.384 & 20 \\
\emph{sn-}  & 0.352 & 16 \\
\emph{scr-} & 0.350 & 20 \\
\emph{br-}  & 0.327 & 22 \\
\emph{sl-}  & 0.324 & 17 \\
\emph{gl-}  & 0.273 & 22 \\
\bottomrule
\end{tabular}
\caption{Mean Purity@1 by phonestheme onset.}
\label{tab:onset_performance}
\end{table}

\paragraph{By suffix.}

\begin{table}[H]
\centering
\begin{tabular}{@{}lcc@{}}
\toprule
Suffix & Mean Purity@1 & $n$ \\
\midrule
\emph{-ling} & 0.452 & 13 \\
\emph{-us}   & 0.445 & 24 \\
\emph{-oid}  & 0.445 & 26 \\
\emph{-a}    & 0.398 & 25 \\
\emph{-or}   & 0.346 & 28 \\
\emph{-ax}   & 0.345 & 31 \\
\emph{-um}   & 0.299 & 28 \\
\emph{-ix}   & 0.295 & 25 \\
\bottomrule
\end{tabular}
\caption{Mean Purity@1 by phonestheme suffix.}
\label{tab:suffix_performance}
\end{table}

The \emph{cr-} onset is the strongest performer---consistent with it being the onset of ``Crungus'' itself. The \emph{-ling} and \emph{-oid} suffixes produce the highest coherence, both carrying strong creature/entity connotations. The \emph{gl-} onset, associated with light and vision rather than creatures or objects, performs worst---visual concepts like ``glow'' may activate diffuse aesthetic associations rather than concrete entities.

\subsection{Contamination Analysis}
\label{sec:contamination}

Rigorous contamination checking was essential. Of the seven phonestheme candidates achieving Purity@1 = 1.0, four were disqualified:

\begin{itemize}
\item \textbf{drudgea:} Visual outputs resembled news/media imagery. ``Drudge'' exists independently (Drudge Report), and the model likely activates associations from that training data rather than constructing from phonesthemes.
\item \textbf{crashor:} While constructed from phonesthemes, ``crash'' is a complete English word, and the agentive \emph{-or} suffix transparently extends it. The model is reading a real word with a suffix, not constructing from sound symbolism.
\item \textbf{skogum:} Outputs showed forest/nature imagery. ``Skog'' means ``forest'' in Swedish and appears in SD~1.5's multilingual training data. The model is translating, not constructing.
\item \textbf{groomus:} Outputs showed wedding/grooming imagery. ``Groom'' is a complete English word; \emph{-us} merely Latinizes it.
\end{itemize}

The two random candidates with Purity@1 = 1.0 were also contaminated: \emph{gopis} (Hindu milkmaids devoted to Krishna, extensively depicted in South Asian art) and \emph{coxen} (Coxen's Fig-Parrot, an endangered Australian bird species).

The single negative control achieving Purity@1 = 1.0 (\emph{wtxmtb}) is likely a CLIP tokenizer artifact---the consonant string may decompose into tokens that happen to activate a coherent region by coincidence. With $n = 50$ negative controls, one spurious hit at the 2\% rate is expected.

The three confirmed cryptids---snudgeoid, crashax, broomix---survived all contamination checks. No real-world entity, product, place, species, or cultural reference matches these strings. Their visual coherence derives from phonesthemic construction alone.

\section{Discussion}
\label{sec:discussion}

\subsection{One Principle, Two Levels}
\label{sec:one_principle}

The central finding of this paper is that morphological pressure creates navigable gradients in text-to-image diffusion models at multiple levels of the generative pipeline.

In Study~1, morphological descriptors at the \emph{training} level---feature descriptions like ``platinum blonde'' and ``beauty mark'' used as LoRA training signal---navigate to a memorized identity basin without requiring the target's name or photographs. The trained LoRA creates a coordinate system that shapes navigation both toward and away from the target, producing structured failure modes (uncanny valley rather than random breakdown) when conditioning pushes into sparse regions.

In Study~2, morphological structure at the \emph{prompt} level---phonesthemic sound patterns like \emph{cr-}, \emph{sn-}, and \emph{-oid} embedded in novel nonsense words---constructs coherent visual entities that never existed in training data. These entities emerge not from semantic content (the words are meaningless) but from the statistical associations that sub-lexical sound patterns carry through the text encoder's tokenization.

The mechanism is the same in both cases. Morphological features---whether visual features (``beauty mark'') or phonological features (\emph{sn-})---activate overlapping regions of the model's latent space. When enough features converge, they address a specific region: an existing identity basin in Study~1, a novel but coherent visual concept in Study~2. The convergence is not random; it follows the statistical structure encoded during training. ``Platinum blonde'' + ``beauty mark'' + ``1950s glamour'' converges on Marilyn because those features co-occurred in her training images. \emph{Cr-} + \emph{-ash-} + \emph{-ax} converges on a dune buggy because impact-words and tool-words and vehicle-words overlap in the training corpus's linguistic statistics.

\subsection{Construction vs.\ Retrieval}
\label{sec:construction_retrieval}

The crungus hunt was designed to answer a specific question: when a nonsense word produces coherent visual output, is the model \emph{retrieving} something it memorized, or \emph{constructing} something from components?

The contamination analysis provides a clear answer: it's both, and the distinction matters.

Four of seven perfect-scoring phonestheme candidates turned out to be retrieving real concepts (drudgea $\to$ Drudge Report, skogum $\to$ Swedish ``forest''). Two random candidates were retrieving obscure real-world referents (gopis, coxen). These are memorization artifacts, not construction. The model recognized a word it had seen during training, even when we thought we had invented it.

But three candidates---snudgeoid, crashax, broomix---are genuine constructions. No training image was ever captioned with these words. No web page contains them. The model built coherent visual concepts from sub-lexical components, the same way a fluent English speaker hearing ``snudgeoid'' would intuit something robotic and sludgy without ever encountering the word before.

This construction capacity has implications for understanding how diffusion models process language. The CLIP text encoder does not treat unknown words as opaque tokens; it decomposes them via BPE into sub-word units and processes the \emph{associations of those units}. The model is, in a meaningful sense, performing morphological analysis---not syntactic parsing of the kind linguists study, but a statistical analog that maps sub-lexical patterns to visual output space.

\subsection{Implications for Latent Space Interpretability}
\label{sec:interpretability}

Both studies reveal that the latent spaces of diffusion models have more navigable structure than typically assumed. The standard view treats latent space as a high-dimensional manifold where meaningful regions are sparsely and unpredictably distributed. Our results suggest a more organized picture:

\begin{itemize}
\item \textbf{Identity basins have sharp boundaries} (Section~\ref{sec:phase_transitions}). Phase transitions between attractors at specific LoRA weights indicate discrete regions separated by decision boundaries, not smooth probability gradients.
\item \textbf{Inverse structure is coherent} (Section~\ref{sec:inverse_navigation}). LoRA training shapes not only the target basin but its geometric complement. The model knows what ``not-Marilyn'' looks like, and that knowledge has structure.
\item \textbf{Sub-lexical patterns map to visual regions} (Study~2). The text-to-visual mapping preserves phonological associations, meaning the text encoder's internal structure is partially interpretable through linguistic analysis.
\end{itemize}

These findings suggest that latent space cartography---systematic mapping of the structure of generative model latent spaces---is feasible using morphological probes at multiple levels.

\subsection{The Phonestheme Gradient}
\label{sec:phonestheme_gradient}

The most surprising quantitative result is not that some phonestheme candidates work perfectly---it's that the \emph{entire distribution} shifts. Phonestheme candidates as a group produce higher Purity@1 scores than random candidates (mean 0.371 vs.\ 0.209). This is not driven by a few outliers; the median also shifts (0.3125 vs.\ 0.0625). Phonesthemic structure provides a general-purpose navigational advantage in visual output space, not just occasional lucky hits.

This gradient suggests that the model's text encoder has internalized something resembling phonestheme sensitivity---an emergent property of training on text corpora where sound-symbolic patterns are statistically present. The text encoder was never explicitly taught that \emph{cr-} means ``impact'' or \emph{-oid} means ``resembling.'' But these patterns exist in English, they appear in training data, and the encoder learned them well enough that they influence visual generation.

\section{Limitations}
\label{sec:limitations}

\textbf{Single model family.} Both studies use Stable Diffusion 1.5 with CLIP ViT-L/14 as the text encoder. We do not know whether the observed effects generalize to other architectures (SDXL, Flux, DALL-E 3), other text encoders (T5, GPT-based), or other training datasets. SD~1.5's specific training corpus (LAION-5B) and tokenizer may produce phonestheme sensitivities that other models lack.

\textbf{Cross-model validation pending.} The three confirmed cryptids (snudgeoid, crashax, broomix) should be tested on other text-to-image models to determine whether their visual coherence reflects universal properties of English phonesthemes or idiosyncrasies of SD~1.5's training.

\textbf{English-centric phonesthemes.} Our phonestheme inventory is drawn entirely from English sound-symbolic patterns. Other languages have different phonestheme inventories (e.g., Japanese \emph{gl-} equivalents, Bantu reduplicated forms). Testing cross-linguistic phonesthemes would clarify whether the effect depends on English-specific training data or more universal sound-symbolic mappings.

\textbf{Limited contamination guarantees.} We checked candidates against English dictionaries, common web corpora, and known cultural references. However, SD~1.5 was trained on LAION-5B, which contains billions of image-text pairs from across the internet. We cannot guarantee that no image in that dataset was captioned with ``snudgeoid'' or ``crashax'' by coincidence. The probability is low---these strings return zero Google results---but non-zero.

\textbf{Purity@1 as a metric.} Purity@1 measures self-clustering in CLIP embedding space, which captures visual similarity as perceived by CLIP. It does not directly measure semantic coherence (whether all outputs depict the ``same thing'' in a way humans would agree on) or novelty (whether the concept is genuinely new vs.\ a recombination of existing concepts). Human evaluation would strengthen the results.

\textbf{Small positive control set.} With only four positive controls (crungus, goblin, fungus, mushroom), the positive control statistics are underpowered. A larger set of known-coherent words would better calibrate the expected distribution.

\textbf{Identity basin scope.} Study~1 uses only Marilyn Monroe as the target identity. While Monroe's deep memorization makes her ideal for demonstrating the principle, we have not tested whether morphological addressing works for less-memorized identities, and the attractor depth threshold for successful navigation remains unknown.

\section{Conclusion}
\label{sec:conclusion}

We have shown that morphological pressure---the statistical associations carried by structured feature descriptions and sub-lexical sound patterns---creates navigable gradients in the latent spaces of text-to-image diffusion models. This principle operates at multiple levels of the generative pipeline.

At the training level, morphological descriptors of constituent features navigate to memorized identity basins through self-distillation, creating LoRA-based coordinate systems that shape both target attraction and inverse repulsion. The resulting geometry has sharp phase transitions between basins, CFG-invariant stability, and structured failure modes that we characterize as ``coherence drag.''

At the prompt level, phonestheme-structured nonsense words produce significantly more coherent visual outputs than random controls ($p < 0.00001$), and three novel words---\emph{snudgeoid}, \emph{crashax}, \emph{broomix}---construct entirely new visual entities from sound symbolism alone. These ``cryptids'' are not retrieved from training data but constructed from the convergent semantic associations of their phonological components.

The existence of these navigational gradients suggests that diffusion model latent spaces are more structured than typically assumed. Morphological probes---whether feature descriptors, phonestheme candidates, or other linguistically motivated constructions---may serve as general-purpose tools for latent space cartography, mapping the geometry of what these models know and how that knowledge is organized.

The original Crungus was a surprise. It emerged from a single prompt and resisted explanation. What this work demonstrates is that Crungus was not an anomaly---it was a signpost. The latent space is full of coherent regions addressable through morphological structure. We need only learn the grammar.

\bibliographystyle{plainnat}
\bibliography{references}

\end{document}